\RequirePackage{fix-cm}
\documentclass[smallextended]{svjour3}       
\smartqed  
\usepackage{graphicx}
\usepackage{mathptmx}
\usepackage{epstopdf}
\usepackage{times}
\usepackage[lined,algonl,boxed]{algorithm2e}
\usepackage{amssymb,amsmath}
\usepackage{multirow}
\usepackage{caption}
\usepackage{subcaption}
\usepackage{paralist}
\usepackage{tikz}
\usepackage{color}
\usepackage{stfloats}
\usepackage{array}
\usepackage{url}
\usepackage{bm}
\usepackage{enumitem}
\usepackage{booktabs, makecell}
\usepackage{natbib}
\usepackage{longtable}
\usepackage[misc,geometry]{ifsym} 

\newcommand{\jiangl}[1]{\textcolor{black}{#1}}

\renewcommand{\cite}[1]{\citep{#1}}
\AtBeginDocument{%
  \paperwidth=\dimexpr
    1in + \oddsidemargin
    + \textwidth
    + 1in + \oddsidemargin
  \relax
  \paperheight=\dimexpr
    1in + \topmargin
    + \headheight + \headsep
    + \textheight
    + 1in + \topmargin
  \relax
  \usepackage[pass]{geometry}\relax
}
\begin{document}
\title{Recent Research Advances on Interactive Machine Learning
}


\author{Liu Jiang        \and
        Shixia Liu \and 
        Changjian Chen
}


\institute{L. Jiang, S. Liu (\Letter), C. Chen \at 
National Engineering Lab for Big Data Software, and School of Software, Tsinghua University \\
              \email{\{jiangl16, ccj17\}@mails.tsinghua.edu.cn,  shixia@tsinghua.edu.cn}      
}

\date{Received: date / Accepted: date}

\maketitle
\begin{abstract}
Interactive Machine Learning (IML) is an iterative learning process that tightly couples a human with a machine learner,
which is widely used by researchers and practitioners to effectively solve a wide variety of real-world application problems.
Although recent years have witnessed the proliferation of IML in the field of visual analytics,
most recent surveys either focus on a specific area of IML or aim to summarize a visualization field that is too generic for IML.
In this paper, we systematically review the recent literature on IML and classify them into a task-oriented taxonomy built by us.
We conclude the survey with a discussion of open challenges and research opportunities that we believe are inspiring for future work in IML.
\end{abstract}
\section{Introduction}
Although machine learning has gained remarkable attention in many fields including information retrieval, computer vision, pattern recognition and data mining.
It is often criticized for its ``black box'' learning, complicated parameter tuning, and lack of human knowledge in the problem solving processing. 
To tackle these issues, interactive machine learning (IML) is emerging as a promising field in visual analytics~\cite{fails2003interactive,liu2017improving,liu2017towards}.
It tightly couples human input with machines in the learning process.
For example, IML can facilitate the performance improvement of machine learning models with the help of human knowledge through rich interactions~\cite{liu2016uncertainty, wang2016topicpanorama, wu2018m3}.
The typical IML workflow includes two steps in each iteration. 
First, the intermediate results of the model as well as the data is visually presented to the user, the user explores the visualization results, draw understandings and insights about the data and the model, then input feedback to the model. 
Second, the model is incrementally updated by integrating the human input.
Hence, IML enables machine learning models to be interactively steered by humans and is more advantageous for the tasks where human knowledge is needed in the analysis process. 

Recent years have seen the surge of IML. 
However, most existing surveys either focus on a specific area of IML, including interactive dimensionality reduction~\cite{sacha2017visual}, 
interactive model analysis~\cite{liu2017towards_survey}, visual social data analytics~\cite{wu2016survey}, and predictive visual analytics~\cite{lu2017state}, or aim to summarize the visualization field that is too generic for IML, e.g., information visualization~\cite{liu2014survey, lu2017frontier}.
The most relevant work to ours is the recent survey by Endert et al.~\cite{endert2017state}, they reviewed visual analytics approaches that integrate machine learning models, with a focus on classification, clustering, regression and dimensionality reduction. 
We elaborate on tasks that are less covered by these existing surveys, such as interactive information retrieval and visual pattern mining.

In this paper, we systematically review recent literature on IML and classify them into a task-oriented taxonomy (Table~\ref{table:tasktaxonomy}) built by us.
We summarize the recent approaches for IML tasks and also point out open challenges and research opportunities that we believe are inspiring for future work in IML.
The purpose of this survey is to provide an overview of the field and help researchers find research topics that are not adequately studied and need further investigation.

\section{Overview and Scope}
\label{sec:overview}
In this survey, we reviewed existing work on IML from visual analytics community, where interactive visualizations empower users to iteratively steer machine learning models.
We reviewed related papers from 2015 to 2018 in the following journals or conferences: IEEE TVCG, IEEE VAST, IEEE InfoVis, EuroVis, and IEEE PacificVis.
To collect the IML papers, we exhaustively checked the titles and abstracts from the proceedings of the conferences (or IEEE TVCG) to find candidates.
Then, we browsed the full texts of uncertain candidates to finalize the selection.
\jiangl{As a result, a total of 99 papers are selected for review.}

As shown in Table~\ref{table:tasktaxonomy}, we build a taxonomy to classify the related papers based on their analysis tasks.
\jiangl{The first-level tasks are first initialized to the ones defined by existing IML surveys on specific areas, e.g., interactive dimensionality reduction  identified by Sacha et al.~\cite{sacha2017visual}.
We then expand them to accommodate papers not pertinent to those initial IML tasks.
For example, we find the visual analytics work on detecting anomaly does not fall into the scope of the initial tasks and thus add ``interactive anomaly detection'' to the first-level IML tasks.}
In total, it results in 9 tasks including classical ones such as ``visual cluster analysis,'' ``interactive dimensionality reduction,'' and recent emerging tasks such as ``interactive model analysis.''
The ``visual cluster analysis'' captures the largest number of the papers \jiangl{(20)} due to the prevalence of data analysis based on interactive clustering.
As high-dimensional data is ubiquitous in practical applications, we see a good number of papers \jiangl{(16)} fall into the task ``interactive dimensionality reduction''.
It is noteworthy that ``interactive model analysis'' also capture a fair amount of papers \jiangl{(14)}.
This is due to the proliferation of recent work on opening the ``black-box'' of deep learning.
\jiangl{It is worth noting that while most papers can be clearly placed to one analysis task (e.g., ``interactive regression''), 
some papers are related to more than one tasks (e.g., ``interactive dimensionality reduction'' and ``visual topic analysis'' often co-occur within one paper).
For the latter cases, we classify papers to the IML tasks that most emphasizes the roles of humans in their particular scenarios.}
\jiangl{To build the taxonomy in the second-level, we follow the previous task taxonomies that cover our first-level tasks.
For example, we refer to ~\cite{liu2018bridging} for categorizing work on ``visual topic analysis'' into ``flat topic analysis,''``hierarchical topic analysis'' and ``topic evolution analysis.''
For the first-level tasks not covered or not being further classified by the previous work, we categorize the related papers based on their focused aspects of a particular first-level task,  e.g., ``visual pattern mining'' is split into ``exploratory event analysis'' and ``mobility pattern analysis.''}

Existing surveys have reviewed related work on part of these tasks.
For example, ``visual cluster analysis,'' ``interactive classification'' and ``interactive regression'' are discussed in~\cite{endert2017state}, ``interactive dimensionality reduction'' is elaborated in~\cite{sacha2017visual}, and ``interactive model analysis'' is presented in~\cite{choo2018visual, hohman2018visual, liu2018visual}.
We focus on tasks less covered by the existing surveys, such as ``interactive information retrieval.''
\begin{small}
\begin{longtable}{p{3cm} p{3.6cm} p{4.2cm}}
\caption{A taxonomy of tasks and the recent representative papers. \jiangl{The bold texts highlight the tasks that we focus on.} The number of papers is shown in the bracket.}\label{table:tasktaxonomy}\\
\renewcommand\arraystretch{1}
\textbf{First-level} & \textbf{Second-level} &  \textbf{Examples} \\ \midrule
 \multirow{13}{*}{\makecell[l]{visual cluster analysis (20)}} & \multirow{1}{*}{\makecell[l]{exploratory data analysis (9)}}&\hspace{1sp}\cite{andrienko2018clustering},~\cite{badam2017steering},~\cite{heimerl2016citerivers},~\cite{purwantiningsih2016visual},~\cite{raidou2016visual},~\cite{sacha2018somflow},~\cite{wu2017mobiseg},~\cite{wu2017streamexplorer},~\cite{xu2016interactive}
\\
\cline {2-3}
 & \multirow{1}{*}{\makecell[l]{comparative \\
 clustering analysis (5)}} & \hspace{1sp}\cite{jarema2015comparative},~\cite{kumpf2018visualizing},~\cite{kwon2018clustervision},~\cite{rieck2016exploring},~\cite{zhang2016visualizing}
\\
\cline {2-3}
 & \multirow{1}{*}{\makecell[l]{bi-cluster analysis (6)}} & \hspace{1sp}\cite{onoue2017quasi},~\cite{sun2016biset},~\cite{watanabe2015biclustering},~\cite{wu2017making},~\cite{wu2015interactive},~\cite{zhao2018bidots}
\\
\midrule
 \multirow{10}{*}{\makecell[l]{interactive dimensionality \\reduction (16)}} & \multirow{1}{*}{\makecell[l]{ subspace analysis (7)}}&\hspace{1sp}\cite{jackle2017pattern},~\cite{liu2015visual},~\cite{wang2018subspace},~\cite{wang2017linear},~\cite{xia2018ldsscanner},~\cite{xie2017visual},~\cite{zhou2016dimension}
\\
\cline {2-3}
 & \multirow{1}{*}{\makecell[l]{high-dimensional \\
data exploration (6)}} & \hspace{1sp}\cite{barbosa2016visualizing},~\cite{berger2017cite2vec},~\cite{cheng2016data},~\cite{jackle2016temporal},~\cite{kwon2017axisketcher},~\cite{liu2016grassmannian},~\cite{stahnke2016probing}
\\
\cline {2-3}
 & \multirow{1}{*}{\makecell[l]{progressive dimensionality\\ reduction (3)}} & \hspace{1sp}\cite{pezzotti2016hierarchical},~\cite{pezzotti2017approximated},~\cite{turkay2017designing}
\\
\midrule
 \multirow{14}{*}{\makecell[l]{interactive model \\analysis (14)}} & \multirow{1}{*}{\makecell[l]{ model understanding (11)}}&\hspace{1sp}\cite{bilal2018convolutional},~\cite{kahng2018activis},~\cite{liu2018analyzing},~\cite{liu2017towards},~\cite{liu2018visual_we},~\cite{ming2017understanding},~\cite{pezzotti2018deepeyes},~\cite{rauber2017visualizing},~\cite{strobelt2018lstmvis},~\cite{wang2018ganviz},~\cite{wongsuphasawat2018visualizing}
\\
\cline {2-3}
 & \multirow{1}{*}{\makecell[l]{model diagnosis (7)}} & \hspace{1sp}\cite{bilal2018convolutional},~\cite{krause2017workflow},~\cite{liu2018analyzing},~\cite{liu2017towards},~\cite{liu2018visual},~\cite{ren2017squares},~\cite{wang2018ganviz}
\\
\midrule
 \multirow{6}{*}{\makecell[l]{interactive classification (8)}} & \multirow{1}{*}{\makecell[l]{ interactive labeling (4)}}&\hspace{1sp}\cite{bernard2018comparing},~\cite{el2017nerex},~\cite{paiva2015approach},~\cite{yu2015iviztrans}
\\
\cline {2-3}
 & \multirow{1}{*}{\makecell[l]{interactive \\feature engineering (2)}} & \hspace{1sp}\cite{brooks2015featureinsight},~\cite{dou2015demographicvis}
\\
\cline {2-3}
 & \multirow{1}{*}{\makecell[l]{parameter spaces analysis (2)}} & \hspace{1sp}\cite{muhlbacher2018treepod},~\cite{rohlig2015supporting}
\\
\midrule
\multirow{6}{*}{\makecell[l]{interactive regression (8)}} & \multirow{1}{*}{\makecell[l]{ interactive \\correlation analysis (5)}}&\hspace{1sp}\cite{klemm20163d},~\cite{shao2017interactive},~\cite{wang2016visual},~\cite{zhang2016visual},~\cite{zhang2015visual}
\\
\cline {2-3}
 & \multirow{1}{*}{\makecell[l]{interactive \\numerical prediction (3)}} & \hspace{1sp}\cite{bryan2015integrating},~\cite{buchmuller2015visual},~\cite{lowe2016visual}
\\
\midrule
\multirow{5}{*}{\makecell[l]{\textbf{interactive}\\ \textbf{information retrieval (8)}}} & \multirow{1}{*}{\makecell[l]{structured information retrieval (2)}}&\hspace{1sp}\cite{pajer2017weightlifter},~\cite{wall2018podium}
\\
\cline {2-3}
 & \multirow{1}{*}{\makecell[l]{unstructured information \\retrieval (6)}} & \hspace{1sp}\cite{behrisch2017magnostics},~\cite{cho2017crystalball},~\cite{liang2017photorecomposer},~\cite{liu2016uncertainty},~\cite{lu2017visual},~\cite{park2018conceptvector}
\\
\midrule
\multirow{5}{*}{\makecell[l]{\textbf{visual pattern mining (8)}}} & \multirow{1}{*}{\makecell[l]{exploratory event analysis (4)}}&\hspace{1sp}\cite{chen2018sequence},~\cite{guo2018eventthread},~\cite{liu2017coreflow},~\cite{liu2017patterns}
\\
\cline {2-3}
 & \multirow{1}{*}{\makecell[l]{mobility pattern analysis (4)}} & \hspace{1sp}\cite{chen2016interactive},~\cite{di2016allaboard},~\cite{krueger2015semantic},
~\cite{poco2015exploring}
\\
\midrule
\multirow{5}{*}{\makecell[l]{\textbf{visual topic analysis (10)}}} 
& \multirow{1}{*}{\makecell[l]{flat topic analysis (4)}}&\hspace{1sp}\cite{alexander2016task},~\cite{el2018progressive},~\cite{glueck2018phenolines},~\cite{thom2015can}
\\
\cline {2-3}
 & \multirow{1}{*}{\makecell[l]{hierarchical topic analysis (4)}} & \hspace{1sp}\cite{kim2017topiclens},~\cite{liu2016online},~\cite{wang2016ideas},~\cite{wang2016topicpanorama}
\\
\cline {2-3}
 & \multirow{1}{*}{\makecell[l]{topic evolution analysis (3)}} & \hspace{1sp}\cite{el2016contovi},~\cite{gad2015themedelta},~\cite{liu2016online}
 \\
\midrule
\multirow{4}{*}{\makecell[l]{\textbf{interactive} \\ \textbf{anomaly detection (7)}}} & \multirow{1}{*}{\makecell[l]{ anomalous points detection (5)}}&\hspace{1sp}\cite{bogl2017cycle},~\cite{cao2018voila},~\cite{leite2018eva},~\cite{lin2017rclens},~\cite{wilkinson2018visualizing}
\\
\cline {2-3}
 & \multirow{1}{*}{\makecell[l]{anomalous sequences detection (2)}} & \hspace{1sp}\cite{cao2016targetvue},~\cite{xu2017vidx}
\\
\bottomrule
\end{longtable}
\end{small}

\section{Analysis on Existing Work}\label{sec:tasks}
In this section, we introduce the recent efforts on IML, with a focus on visual pattern mining, interactive anomaly detection, interactive information retrieval, and visual topic analysis.
\jiangl{We briefly discuss other tasks elaborated by existing IML surveys.}
\begin{figure*}[!tb]
\centering
    \begin{subfigure}{0.53\textwidth}
          \includegraphics[width=\textwidth]{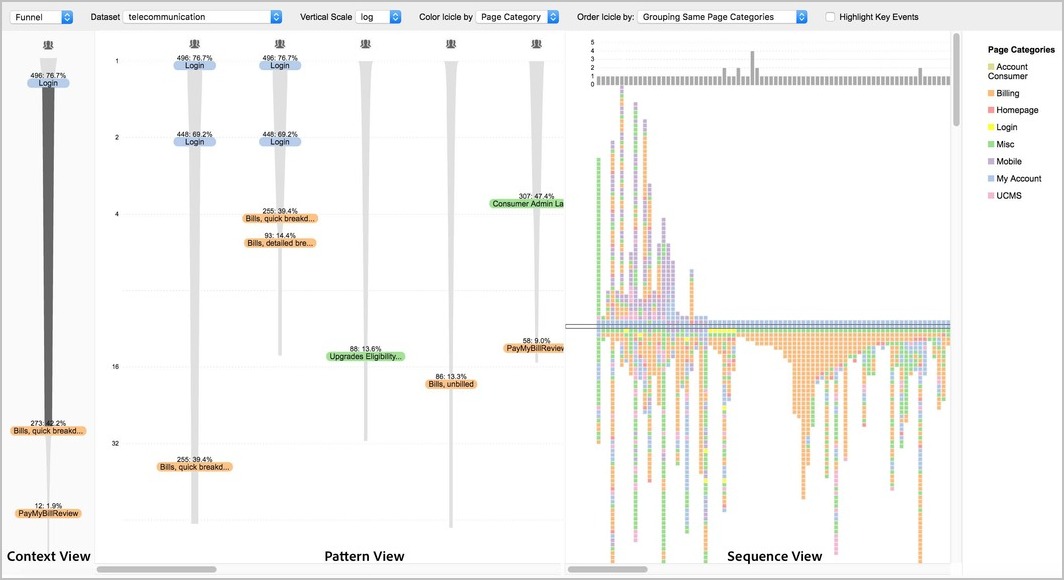}
    \caption{}\label{fig:sequence_clickstream}
    \end{subfigure}\hspace{8mm}
\centering
    \begin{subfigure}{0.38\textwidth}          \includegraphics[width=\textwidth]{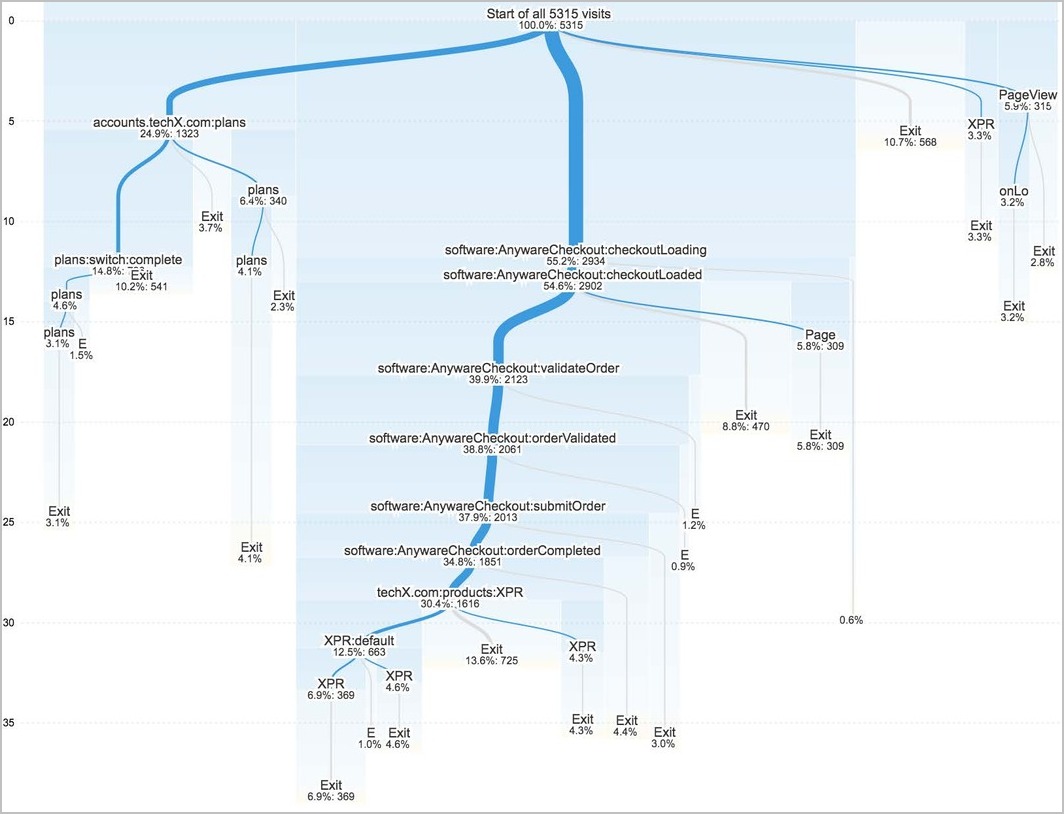}
    \caption{}\label{fig:sequence_coreflow}
	\end{subfigure}
    \caption{Examples of visual pattern mining for exploratory event analysis: (a) The level-of-detail visualizations of web clickstreams~\cite{liu2017patterns}: the patterns and segments are visualized using a funnel metaphor (context and pattern view), with their details such as sequences and events shown in an icicle-like plot (sequence view);
(b) CoreFlow~\cite{liu2017coreflow}: the branching pattern is visualized in a node-link diagram with an icicle plot as its background to indicate the length and amount of event sequences.}
\end{figure*}
\subsection{Visual Pattern Mining}
Visual pattern mining is widely used to interactively discover latent patterns for exploratory correlation or association analysis~\cite{liu2017coreflow,lu2016exploring,poco2015exploring,wen2010mining,yan2018visual}.
Recent work on visual pattern mining focuses on effectively tackling data-driven tasks including understanding user travel patterns from social media~\cite{chen2016interactive}, analyzing customer behaviors from web clickstreams~\cite{liu2017coreflow, liu2017patterns} and exploring traffic dynamics to improve urban infrastructure~\cite{poco2015exploring}.
They fall into two categories: exploratory event analysis~\cite{chen2018sequence, guo2018eventthread, liu2017coreflow, liu2017patterns} and mobility pattern analysis~\cite{chen2016interactive, di2016allaboard, krueger2015semantic, poco2015exploring}.
\subsubsection{Exploratory Event Analysis}
Event sequences, i.e., a collection of series that each represents timestamped event occurrences~\cite{chen2018sequence, guo2018eventthread}, can model many real-world data such as user interaction log, machine alert messages and vehicle maintenance records.
Exploratory event analysis mainly focuses on discovering the overall sequential patterns as well as investigating relationship (e.g., causality) between individual events.
Recent work can be classified into two categories: frequency-based~\cite{ liu2017coreflow, liu2017patterns} and clustering-based~\cite{chen2018sequence, guo2018eventthread}.

For frequency-based approaches, frequent pattern mining algorithms are often adopted to extract sub-sequences that frequently occur in the data.
For example, Liu et al.~\cite{liu2017patterns} adopt a frequency-based algorithm to mine maximal sequential patterns from web clickstreams for understanding the shared browsing path of customers.
To enable a visual exploration, the mined patterns along with their supporting sequences and events are visualized in a level-of-detail manner, based on a funnel visual metaphor and an icicle-like plot (Fig.~\ref{fig:sequence_clickstream}).
Rich interactions are provided to facilitate the exploration, such as aligning events for contextual comparison, and drilling down in a selected pattern segment to disclose its finer-level structures.
Noticing that the traditional frequency-based approach does not scale to a large-scale dataset and the relationships between the mined patterns remain unclear, 
Liu et al.~\cite{liu2017coreflow} later propose a novel rank-divide-trim algorithm to extract a branching pattern from the event sequences.
The branching pattern is more interpretable to users since it reveals the relationships of frequent patterns in a unified view.
To present the branching pattern, the icicle plot and node-link diagrams are combined to visualize its tree structure as well as the statistics associated with the branches (Fig.~\ref{fig:sequence_coreflow}).
To capture users' interests in specific events, they allow users to define high priority events to inform the mining algorithm to guarantee their appearance in the  pattern. 

For clustering-based approaches, researchers mainly use sequence clustering algorithms to summarize the data for pattern analysis.
For example, Guo et al.~\cite{guo2018eventthread} formulate the event sequences as a three-way tensor, the dimensions of which correspond to entity, stage, event, respectively.
Based on the tensor decomposition, they cluster event sequences to threads that each represent a latent sequential pattern.
The threads are visualized using a line map visual metaphor, and can be interactively grouped by users to reveal the temporal evolution of patterns, such as merging and splitting.
Chen et al.~\cite{chen2018sequence} propose an information-theoretic approach for summarizing the event sequences.
Specifically, they follow the minimum description length principle~\cite{grunwald2007minimum} to find a summarization that balances the conciseness and the information loss.
Based on this approach, they group the data to a set of sequential patterns that stand as a summarization of data.
Due to the lossy nature of the summarization, 
the correction needed to recover the data from the summarization is visualized along with the patterns for a more faithful data representation. 
\begin{figure}[!tb]\centering
\includegraphics[width=0.90\linewidth]{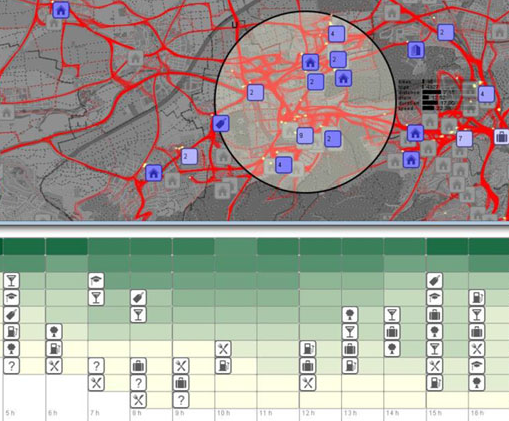}
\caption{Movement patterns semantically enriched with POIs~\cite{krueger2015semantic}: the geographic view highlights the frequent routes and the aggregated POIs with uncertainty encoded as color intensity. The frequently visited POIs in different time are shown in the heatmap.}\label{fig:move_enrich_POI}
\end{figure}
\subsubsection{Mobility Pattern Analysis}
Recent work mainly focuses on analyzing mobility patterns in geospatial movement sequences including traffic trajectories~\cite{krueger2015semantic, poco2015exploring}, cellphone mobility records~\cite{di2016allaboard} and geo-tagged microblogs~\cite{chen2016interactive}.
To capture the geospatial characteristics of movement sequences, researchers often employ geographic maps to visualize the mobility patterns.

Lorenzo et al.~\cite{di2016allaboard} present AllBoard, a visual analytics approach for exploring user mobility from cellphone data.
Specifically, they employ a pattern-growth-based algorithm to extract the shared route patterns, which are mapped to the urban transport network to visualize their geographical topology.
Examining the route patterns, users can explore potential directions for optimizing public transport.
Poco et al.~\cite{poco2015exploring} propose a closest path traffic model to calculate the taxi trips and traffic speed from the sparse traffic data.
Using vector field visualizations, the taxi trips along with the traffic speed are visualized in an animated manner, which helps users understand the dynamic traffic pattern of the urban infrastructure.

While those work does not address the inherent uncertainty associated with the data,  modeling and tackling uncertainty with human-in-the-loop can establish a more reliable mobility analysis.
Kruger et al.~\cite{krueger2015semantic} present a visual analytics approach for semantic movement analysis.
They enrich the geospatial movements with POIs obtained from social media.
The POIs are aggregated in the geographical map to indicate the semantics of movements, and the frequent POIs of a focused region are visualized in a heatmap to show the temporal dynamics (Fig.~\ref{fig:move_enrich_POI}).
As the imprecision of the measurement on time and space can introduce uncertainty in assigning a POI to a geographic position, they propose a mixed model to quantify the uncertainty.
For positions with highly uncertain POIs, users can explore their geospatial contexts and determine their POI categories to resolve the uncertainty.
Chen et al.~\cite{chen2016interactive} develop a visual analytics system for exploring Weibo user movement patterns and their semantics.
In particular, they employ Gaussian mixtures to model the uncertainty of movement time interval caused by the sparsity of geo-tagged Weibo data.
The output of Gaussian mixtures are visualized in line charts, where users can interactively steer the model by tuning the number of mixtures and filter the data to reduce the uncertainty.
The filtered data is used to extract reliable movement patterns that are aggregated in a geographic map with the contextual Weibo messages displayed for a semantic exploration.

\begin{figure}[!tb]
  \centering
    \includegraphics[width=0.70\linewidth]{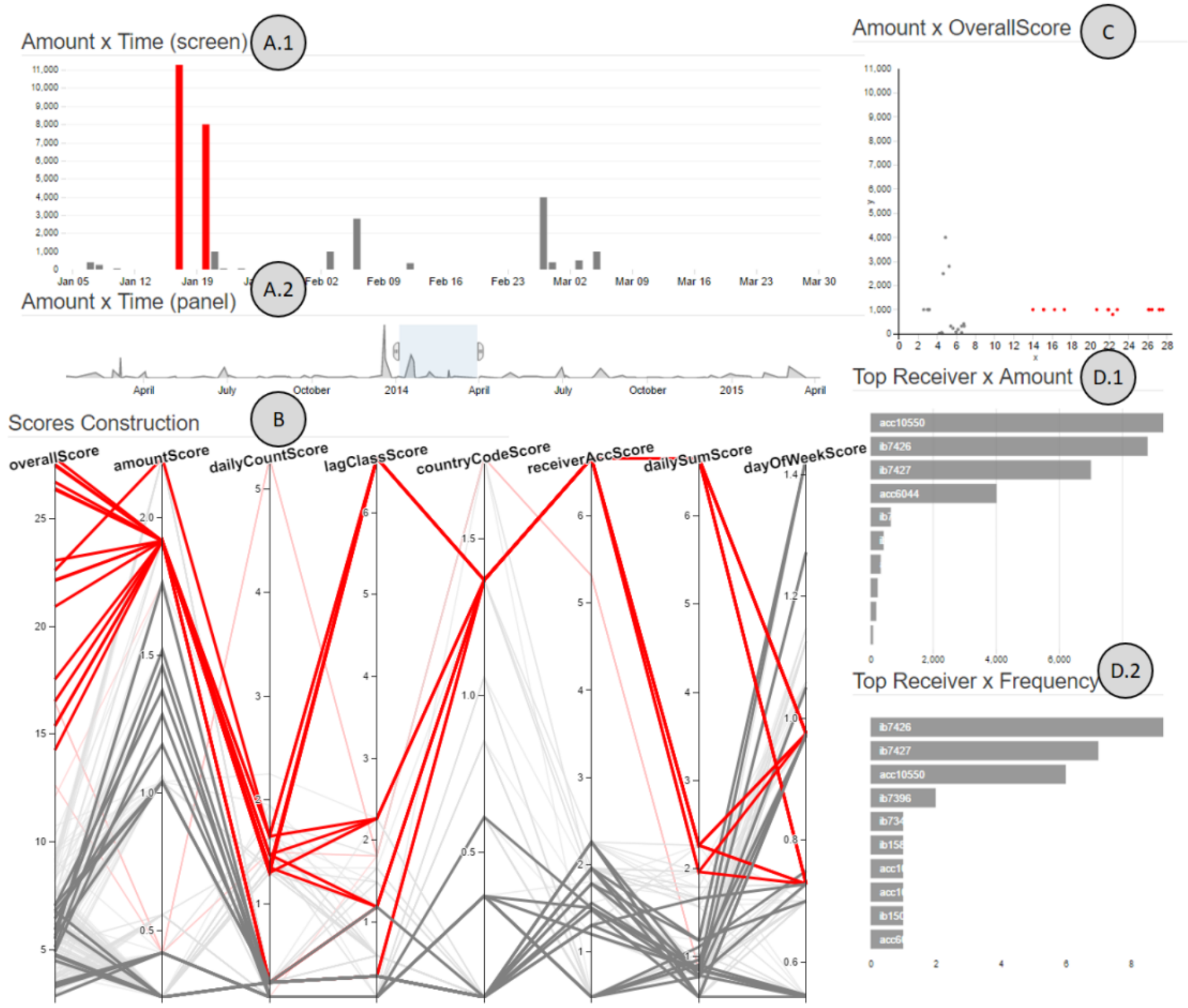}
    \caption{EVA~\cite{leite2018eva}: the parallel coordinate plot shows the contribution of each sub-score to the overall score. The detected anomalous transactions are highlighted.}
    \label{fig:AnomalyDetection_points}
\end{figure}
\begin{figure*}[!tb]\centering
\includegraphics[width=\linewidth]{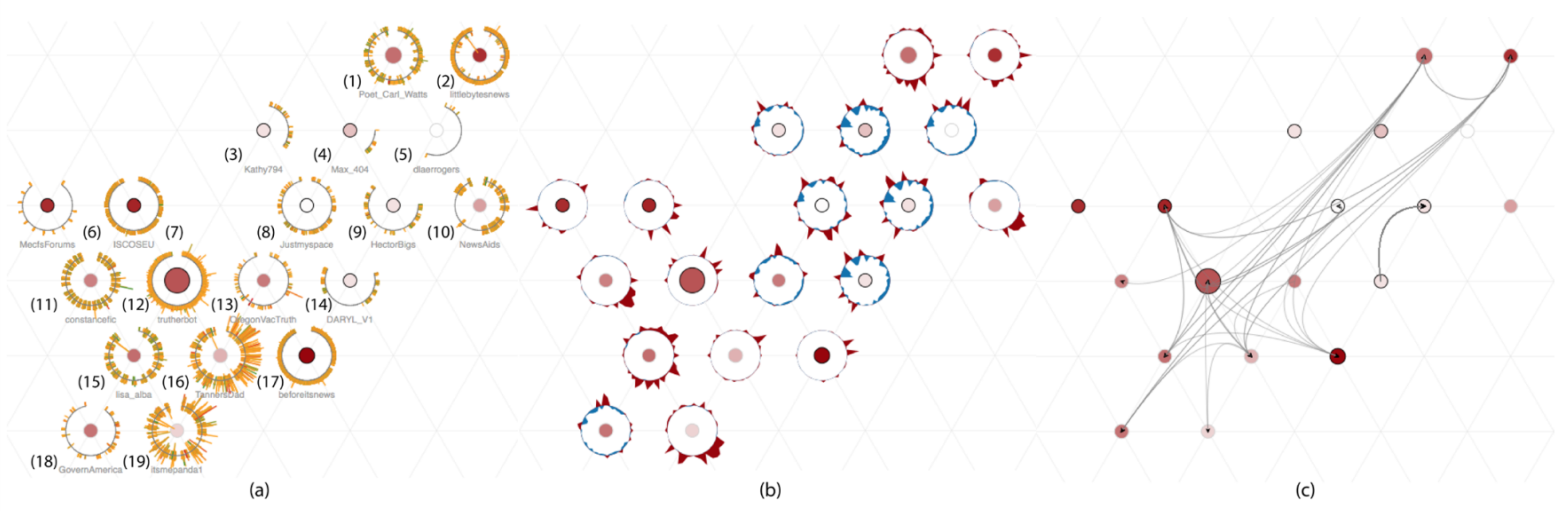}
\caption{TargetVue~\cite{cao2016targetvue}: (a) behavior glyph shows the activities of a user over time; (b) z-glyph shows z-score of features; (c) relation glyph shows relationships of users in online communication systems.}\label{fig:AnomalyDetection_sequences}
\end{figure*}
\subsection{Interactive Anomaly Detection}

Interactive anomaly detection refers to the disclosure of anomalies, which are defined as the data with unexpected behaviors~\cite{han2011data}, with human in the process. 
We classify the recent literature into two categories: anomalous points detection ~\cite{bogl2017cycle, cao2018voila, leite2018eva, lin2017rclens, wilkinson2018visualizing} and anomalous sequences detection~\cite{cao2016targetvue, xu2017vidx}. 

\subsubsection{Anomalous Points Detection}
Anomalous points detection refers to the detection of individual points with unexpected behaviors.
For example, Leita et al. propose EVA~\cite{leite2018eva}, a visual analytics system for interactive fraudulent transaction detection.
They employ a profile-based algorithm to characterize customers based on their historical transactions.
To measure the overall anomaly score of an incoming transaction, a scoring system is used to compare the transaction with the profile of its corresponding customer.
Specifically, the overall anomaly score is calculated by summarizing the sub-scores of the transaction, which are visualized in a parallel coordinate plot (Fig.~\ref{fig:AnomalyDetection_points}B) to show their contribution to the overall score.
Exploring the parallel coordinate plot, users can relate each sub-score to the overall score for comparatively analyzing the pattern of an anomalous transaction.
Anomalous points detection are more challenging for streaming spatio-temporal data due to its massive volume and heterogeneity~\cite{cao2018voila} .
To address this issue, Cao et al. propose Voila~\cite{cao2018voila}, a visual analytics approach for online detecting anomaly on streaming spatio-temporal data.
Specifically, they formulate the spatio-temporal data as a tensor and decompose it to derive latent representations of region, time and feature.
Thus, the anomalous patterns can be extracted from incoming data by comparing the incrementally updated latent representations to that in history.
They use geographic visualization to show the anomalous patterns, which enables a user to provide feedback such as confirming or rejecting an anomalous pattern, to interactively steer the detection model.

\subsubsection{Anomalous Sequences Detection}
Anomalous sequences detection is the process of identifying sequences with unexpected behaviors.
Cao et al.~\cite{cao2016targetvue} propose TargetVue to detect Internet users 
with anomalous behaviors in online communication systems. 
To characterize the behaviors of Internet users, they extract a set of features from multiple perspectives, such as the user post and interaction between users. 
They visualize the characterized behaviors using glyphs (Fig.~\ref{fig:AnomalyDetection_sequences}), which enables a user to detect and reason about the anomalous patterns.
Another interesting application of anomalous sequences detection is the performance diagnostics of assembly line.
Assembly line data contains a massive amount of logs collected from the product manufacturing process.
To address the diagnostics on assembly line, Xu et al.~\cite{xu2017vidx} propose a visual analytics approach, ViDX, based on adapting the Marey's graph to represent each process as a polyline. 
To reduce the visual clutter caused by massive amount of data, ploylines representing normal processes are aggregated as thick bands while anomalous ones are highlighted. 
The timeline of the Marey's graph is the topological sort of the abstract directed acyclic graph representation of assembly line, which facilitates users to trace the product history.
By interactively incorporating user feedback, the user-identified
anomalous processes are propagated to other processes for improving the detection accuracy.


\begin{figure*}[!tb]
\centering
    \begin{subfigure}{0.56\textwidth}
          \includegraphics[width=\textwidth]{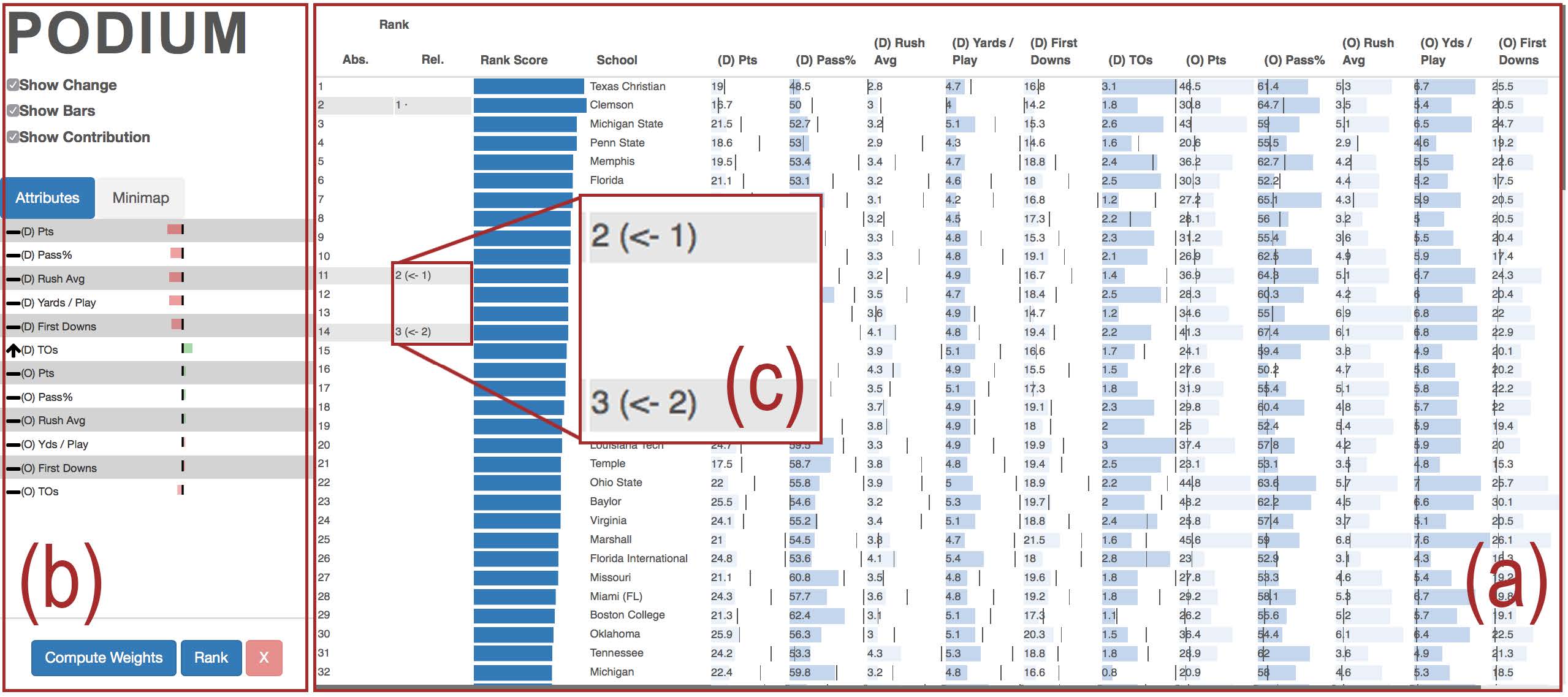}
    \caption{}\label{fig:podium}
    \end{subfigure}
\centering
    \begin{subfigure}{0.43\textwidth}
          \includegraphics[width=\textwidth]{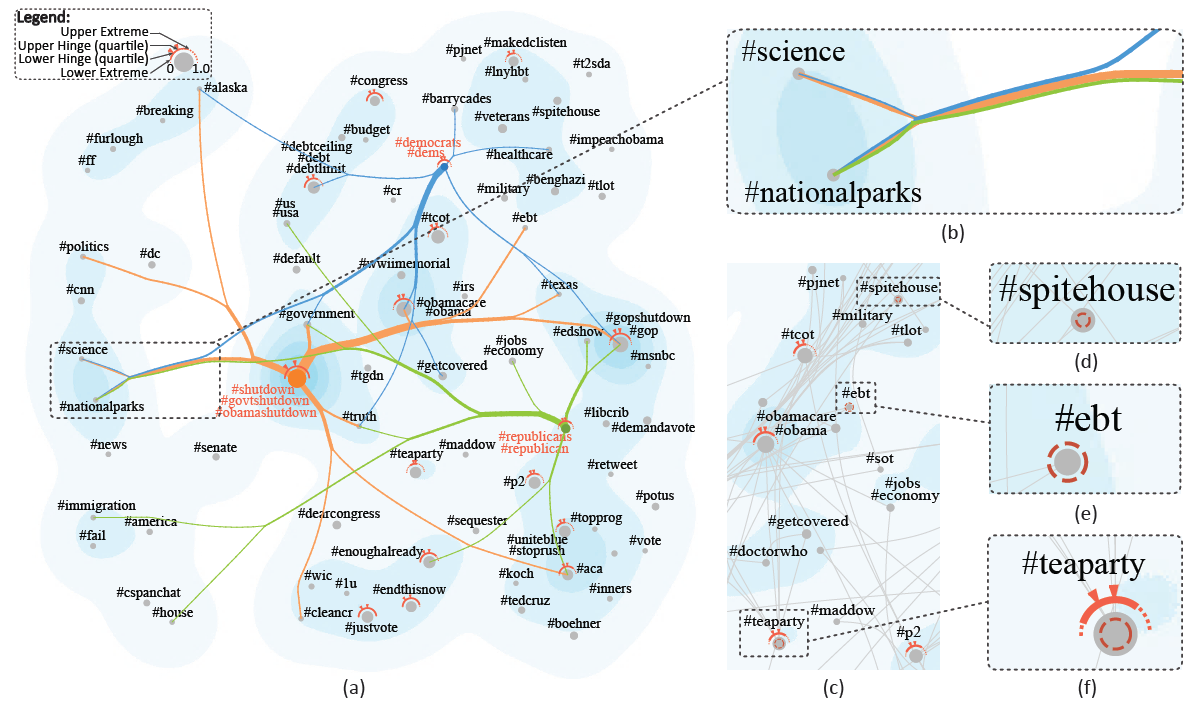}
    \caption{}\label{fig:mutual_ranker}
	\end{subfigure}
    \caption{Example work on interactive information retrieval: (a) Podium~\cite{wall2018podium}: the table view represents each instance as a row and its attributes as columns. Users can drag instances to rank a subset of data; (b) MutualRanker~\cite{liu2016uncertainty}: the composite visualization contains the graph visualization, flow map and glyph,  which visualize ranking results, uncertainty propagation and uncertainty distribution, respectively.}
\end{figure*}
\subsection{Interactive Information Retrieval}
We refer to interactive information retrieval as visually ranking or recommending instances for returning relevant information to a query or interactively finding optimal solutions for a decision making.
It has been recently studied in many data-driven applications, such as microblog retrieval~\cite{liu2016uncertainty}, network exploration~\cite{behrisch2017magnostics}, event analysis~\cite{lu2017visual,cho2017crystalball}, document analysis~\cite{park2018conceptvector, wei2010irank} and evolutionary ranking analysis~\cite{lei2016visual}.
Recent work falls into two categories: structured and unstructured information retrieval.
\subsubsection{Structured Information Retrieval}
Structured information retrieval is widely used to rank the multi-attribute tabular data to support multi-criteria decision making.
As the ranking results are dependent on the weight of each attribute, recent work mainly focus on exploring the weight space~\cite{pajer2017weightlifter} and interactively learning the weights based on user preferences~\cite{wall2018podium}.

Pajer et al. develop WeightLifter~\cite{pajer2017weightlifter}, a visual analytics system that enables users to visually explore the attribute weight space and gain an understanding of the sensitivity of the ranking to the changes of weights.
Their main idea is to visualize slices of weight space and enrich the slices with the projected sensitivity information.
Specifically, they use a vertical axis and barycentric coordinates to show the 1D and 2D slices, respectively.
To visualize the sensitivity of ranking, they segment the weight space based on MCMC sampling and project the segments to the 1D or 2D slices.
By exploring the slices, users can examine the regions where the ranking results are (in)sensitive to changes of weights, and interactively impose constraints on weight space to filter out implausible ranking results.

As the data volume and dimensions grow, it may be cognitively hard for humans to specify weights of attributes for ranking the whole data.
Thus, a more plausible way is perhaps to use machine learning models to learn the weights from users' preference. 
Motivated by this, Wall et al. propose Podium~\cite{wall2018podium} to interactively rank the multi-attribute data based on user preference.
They employ Ranking SVM to infer the weight of each attribute from the user preference (i.e., the ranking of a subset of the data).
Particularly, the multi-attribute data are visualized in a table view (Fig.~\ref{fig:podium}), where users can drag the instances to communicate his preference.
\subsubsection{Unstructured Information Retrieval}
Interactive information retrieval are recently applied to unstructured data such as texts from social media~\cite{lu2017visual, liu2016uncertainty, cho2017crystalball}, newspaper article comments~\cite{park2018conceptvector}, matrices collections~\cite{behrisch2017magnostics} and photos~\cite{liang2017photorecomposer}.
Due to the unstructured nature of data, characterizing data instances and capturing their relationships are not straightforward and are often specific to data types and applications.
For example, while cosine distance is widely used to measure the similarity between TFIDF features of posts in microblog retrieval~\cite{liu2016uncertainty}, the word embeddings are often employed to characterize words and documents in semantic retrieval applications~\cite{park2018conceptvector}.

Liu et al.~\cite{liu2016uncertainty} propose MutualRanker to interactively retrieve salient data from microblogs and address the uncertainty introduced by the ranking model.
Specifically, they derive the relationships within and between posts, users and hashtags, using the content information of posts as well as their metadata such as hashtag co-occurrences.
Based on the derived relationships, they employ a mutual reinforcement graph model~\cite{duan2012twitter} to rank posts, users and hashtags,
as well as modeling the ranking uncertainty using Poisson mixtures.
The uncertainty of ranking results and its propagation is revealed in a composite visualization (Fig.~\ref{fig:mutual_ranker}), which enables users to refine the ranking results to tackle the uncertainty.

As semantic similarities between words are important for retrieving documents to a given topic or concept, recent work introduces human-in-the-loop model to interactively learn these semantic similarities.
For example, to retrieve documents relevant to a given topic, Lu et al.~\cite{lu2017visual} propose a user-guided matching model that loops 
human in calculating semantic similarity between words in documents and that in a topic.
Specifically, semantic similarities of words are initially calculated by a knowledge-based method and users can modify the similarities through semantic interaction, i.e., dragging a word from one cluster to another in the force-directed layout.
Consequently, their semantic similarities are adjusted and the relevance of documents are updated.
Along the same line, Park et al.~\cite{park2018conceptvector} present ConceptVector, an interactive concept building system for document analysis.
Users specify initial seed keywords to retrieve more relevant words based on the semantic similarities derived from their word embeddings.
The retrieved words are projected in a t-SNE plot, which enables users to visually explore their similarities, label them as relevant or irrelevant.
By modifying related keywords of a concept, users can interactively retrieve the relevant documents for a finer-level analysis.

Cho et al.~\cite{cho2017crystalball} presents CrystalBall for visually analyzing future events extracted from social media.
To identify future events, they propose seven measures to characterize the events and adopt RankSVM~\cite{joachims2002optimizing} to evaluate their quality.
To present the high quality future events, interactive visualizations are developed to display the ``who, what, when, where'' information, which allows users to visually retrieve the events of interests.

There is some initial work on applying information retrieval on matrix data and photos.
Behrisch et al.~\cite{behrisch2017magnostics} address the task of matrix retrieval based on the feature descriptors of visual patterns such as diagonal blocks.
Specifically, they rank matrices based on the Euclidean distances between their feature descriptors to that of a query matrix.
To select informative feature descriptors, they evaluate each descriptor by the proposed criteria such as pattern discrimination.
Their matrix retrieval approach can be integrated in visual analytics systems to interactively retrieve the matrix and guide the exploration.
Liang et al.~\cite{liang2017photorecomposer} propose PhotoRecomposer, an interactive photo recomposition method based on learning the composition distance metric from user preference.
By ranking the example photos using the learned metric, several reference photos are recommended for recomposing the user photos. 

\begin{figure*}[!tb]
\centering
    \begin{subfigure}{0.44\textwidth}
          \includegraphics[width=\textwidth]{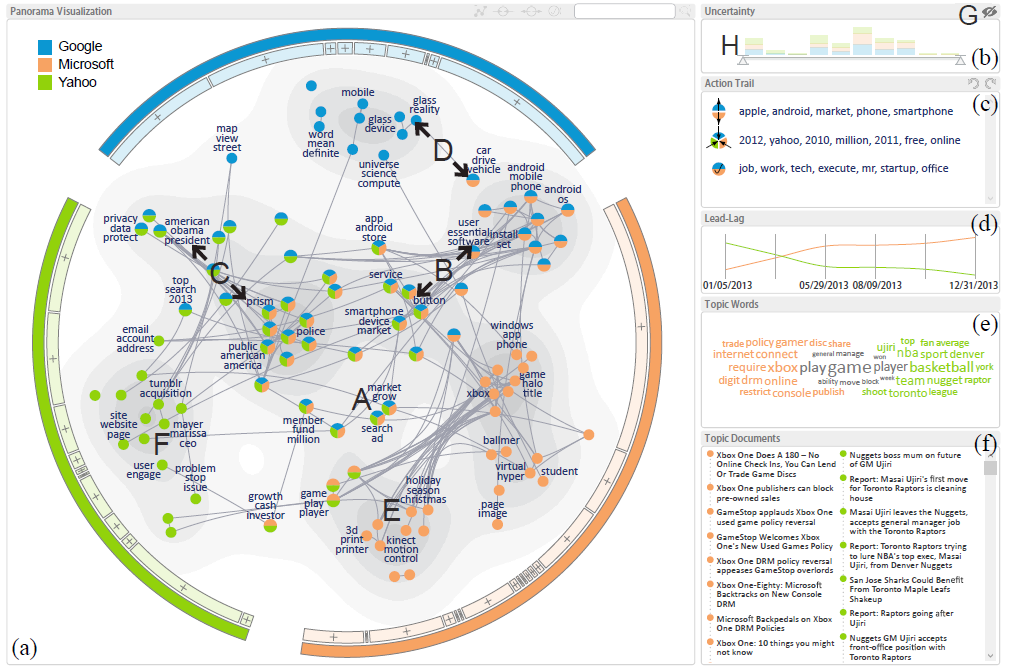}
    \caption{}\label{fig:topic_panorama}
    \end{subfigure}
\centering
    \begin{subfigure}{0.55\textwidth}
          \includegraphics[width=\textwidth]{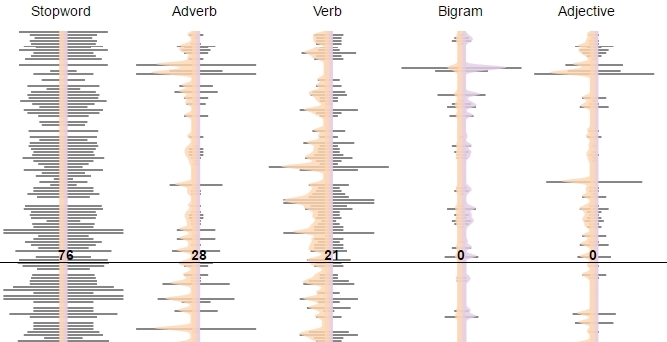}
    \caption{}\label{fig:violin_plots}
	\end{subfigure}
    \caption{Example work on visual topic analysis: (a) TopicPanorama~\cite{wang2016topicpanorama}: the graph visualization shows a full picture of topics across multiple sources. The radial icicle plot visualizes the topic hierarchy around the circumference; (b) Violin plots~\cite{el2018progressive} comparatively visualize the weight distribution of each word class from two topic models (encoded by colors).}
\end{figure*}
\subsection{Visual Topic Analysis}
Visual topic analysis is a popular visual analytics technique for understanding and analyzing latent topics inside a document collection.
Each topic is represented by a set of keywords derived from the documents and each keyword has a probability value to indicate its relevance to that topic.
Recent efforts on visual topic analysis focus on efficiently handling  real-world tasks such as analyzing topics from multiple sources~\cite{jiang2016text,wang2016topicpanorama}, characterizing disease subtypes from clinical  data~\cite{glueck2018phenolines} and review analysis~\cite{xu2018vaut}.
Following the task taxonomy of a recent text visualization survey~\cite{liu2018bridging}, we summarize the work pertinent to visual topic analysis from three perspectives: flat topic analysis, hierarchical topic analysis and topic evolution analysis.
\subsubsection{Flat Topic Analysis}
Recent work on analyzing topics organized in a flat structure mainly addresses topic comparison task.
They can be classified into two categories: within-model topic comparison, between-model topic comparison.

The within-model topic comparison can disclose how topics are related to each other, which facilitates users to examine semantic differences of topics and make a more informed optimization~\cite{glueck2018phenolines}.
Recent works mainly display the juxtaposition of topic visualizations for comparing different topics.
For example,  Glueck et al.~\cite{glueck2018phenolines} presents PhenoLines, a visual analysis tool for disease subtype analysis based on topic modeling.
They apply LDA~\cite{blei2003latent} on electronic health care records, where patients, phenotypes and disease subtypes are modeled as documents, words and topics, respectively.
Mapping the words to a phenotype ontology, each topic is visualized in a sunburst diagram.
They juxtapose the sunburst diagrams of different topics to enable a between-topic comparison.
Some actionable insights are derived from the comparison. 
For example, they show that a prior distribution that emphasize the words that frequently occur in the data, can better differentiate topics than a non-informative Dirichlet prior.
Along the same line, Thom et al.~\cite{thom2015can} juxtapose the word clouds that visualize the keywords of topics, to enable a semantic comparison between topics for situational awareness analysis on social media.

The between-model topic comparison can help validate the gained insights~\cite{alexander2016task}, select models and tune their parameters~\cite{alexander2016task, el2018progressive}.
Alexander and Gleicher~\cite{alexander2016task} propose a task-driven approach for between-model topic comparison.
They characterized three main specific comparison tasks: topic alignment, distance comparison and timeline comparison.
To tackle these tasks, they apply existing visualization techniques such as heatmaps and parallel coordinates, as well as a novel ``buddy plot'' for examining the consistency of document distances derived from two topic models.
El-Assady et al.~\cite{el2018progressive} propose a progressive visual analytics approach to optimize weights of word class (e.g., nouns).
They present comparative visualizations of two topic modeling results for users to examine topic matching, explore topic semantics, understand distributions of weights over documents, and inspect the texts of documents for communicating his or her preference of one topic model.
In particular, they propose ``violin plots'' to enhance the comparison between weight distributions from two topic models (Fig.~\ref{fig:violin_plots}).
The preference of the user is used to adjust the weights of the affected keywords, i.e., the weight of keywords from the favored topic model is increased while the keyword from the other model is penalized.
\subsubsection{Hierarchical Topic Analysis}
Organizing the topics into a hierarchy can alleviate the visual clutter problem caused by the massive amount of topics, and provide richer information to reveal the latent relationships between topics~
\cite{dou2016topic, wang2016topicpanorama}.
Recently, researchers develop level-of-detail visualizations to present the hierarchical topic relationships to users,  allow them to dive into the details and refine the topic hierarchy.

Wang et al.~\cite{wang2016topicpanorama} present TopicPanorama, a visual analytics system for understanding the topics across multiple sources.
They extract topic graphs from multiple sources using correlated topic models~\cite{chen2013scalable}, then match them using a consistent graph matching algorithm and cluster them into the topic hierarchies by a constraint-based algorithm.
The matching results are presented in a graph visualization with
topic hierarchies displayed by a radial icicle plot around the circumference (Fig.~\ref{fig:topic_panorama}), which enables users to find common topics discussed across multiple sources and distinctive ones covered in each individual source.
By incorporating an online metric learning algorithm, users are allowed to interactively modify the graph matching results based on their knowledge.
There is some recent work on refining topic hierarchy based on altering the cluster number.
Kim et al.~\cite{kim2017topiclens} propose a nonnegative-matrix-decomposition-based topic model that can build the topic hierarchy in a user-driven manner.
The initial topic hierarchy is learned by the topic model with a predefined cluster number.
Users can refine the initial hierarchy by overlaying a lens on a region of interest in the t-SNE plot of documents, to trigger an increase of the cluster number.
Consequently, the nodes of the hierarchy that accommodate the documents in the focused region, are split into more clusters, which reveals a finer-level structure of topics.
\subsubsection{Topic Evolution Analysis}
Analysis of topic evolution is important to understanding dynamic patterns in time-varying documents~\cite{liu2014survey, liu2016online}.
For example, Gad et al.~\cite{gad2015themedelta} quantify the temporal shift of topics by measuring the keyword overlap between topics.
Based on the shift of topics, they propose a temporal segmentation algorithm to detect the time step where the topic distributions are significantly changed.
By visualizing the common keywords across time segments using trendlines, the temporal dynamics of topics such as merging or splitting are disclosed.
El-Assady et al.~\cite{el2016contovi} develop ConToVi, a visual analytics system for exploring thematic dynamics of speaker utterances in multi-party conversation.
Specifically, they employ a hierarchical topic model to extract the topic structures and the speaker-topic relationships, which are visualized in a topic-space radial layout.
Comparing the positions of speakers in the topic-space, users can explore the thematic focuses of speakers.
By animating the radial layout, users can track the topic shift indicated by the transition of speaker positions.
\subsection{Discussion on Other Tasks}\label{sec:discuss_others}
\noindent\textbf{Visual cluster analysis}. \jiangl{We refer to visual cluster analysis as the task of interactively partitioning instances to multiple groups based on their similarities.
It uncovers grouping patterns in the data and stands as a fundamental approach for visually aggregating or summarizing the data.
Thus, visual cluster analysis has been applied to many data-driven applications such as weather ensemble forecast~\cite{kumpf2018visualizing} and air traffic optimization~\cite{andrienko2018clustering}.
Recent work mainly focuses on cluster-based exploratory data analysis~\cite{andrienko2018clustering, badam2017steering, heimerl2016citerivers, sacha2018somflow, wu2017mobiseg, wu2017streamexplorer}, comparative clustering analysis~\cite{jarema2015comparative, kumpf2018visualizing, kwon2018clustervision, zhang2016visualizing}, and bi-cluster analysis~\cite{sun2016biset, watanabe2015biclustering, wu2017making, wu2015interactive, zhao2018bidots}.}
\\

\noindent\textbf{Interactive dimensionality reduction}. \jiangl{Recent years have seen a plethora of work on interactive dimensionality reduction for visually analyzing high-dimensional data.
They mainly address subspace analysis~\cite{jackle2017pattern, liu2015visual, wang2018subspace, wang2017linear} and progressive high-dimensional data exploration~\cite{pezzotti2016hierarchical, pezzotti2017approximated, turkay2017designing}.
One important component in human-centered dimensionality reduction is the interaction.
They can be classified into three categories: manipulating data points (e.g., dragging and dropping instances to define constraints), interacting with features (e.g., tuning weights of dimensions) and modifying model variables (e.g., rotating projection bases).
For more details about the interaction paradigm, we refer readers to existing surveys such as~\cite{endert2017state, sacha2017visual} that have comprehensively categorized and analyzed the interaction intent and scenarios.}
\\

\noindent\textbf{Interactive classification}. \jiangl{Recently, interactive classification has been applied to a wide variety of applications including drug relation analysis~\cite{verma2017drugpathseeker}, urban planning~\cite{yu2015iviztrans}, relationship exploration in multi-party conversations~\cite{el2017nerex} and user demographic analysis~\cite{dou2015demographicvis}.
The quality of labels, features and parameter tuning are important factors to building a high-performance classifier.
Hence, based on which factor the interactive classification focuses on, we classify recent work into three categories: interactive labeling, interactive feature engineering and parameter space analysis.
The work on interactive labeling mainly loop humans in removing annotation noise and inspecting the labels of the most uncertain instances~\cite{bernard2018comparing}.
For interactive feature engineering, recent work focuses on visually constructing discriminative features~\cite{brooks2015featureinsight} and facilitating users to understand the correlation between features and classes~\cite{dou2015demographicvis}.
Recent work related to parameter space analysis mainly propose interactive visualization for users to navigate and explore parameter space~\cite{rohlig2015supporting} and provide visual cues to users for understanding the influence of the parameters~\cite{muhlbacher2018treepod}, such as the sensitivity of the classification results to parameters.}
\\

\noindent\textbf{Interactive regression}.
\jiangl{While interactive classification deals with categorical labels, interactive regression refers to the prediction of continuous values by modeling the relations between independent and dependent variables.
Recent work applies interactive regression to many real-world problems such as disease risk factor analysis~\cite{klemm20163d}, causality reasoning~\cite{wang2016visual} and prediction of scientific emulation~\cite{bryan2015integrating}.
They can be categorized into interactive correlation analysis~\cite{klemm20163d, shao2017interactive, wang2016visual, zhang2016visual, zhang2015visual} and interactive numerical prediction~\cite{bryan2015integrating, buchmuller2015visual, lowe2016visual}.
Interactive correlation analysis focuses on visually mining relationships among variables of data, based on the regression models.
Users can inject their domain knowledge to the regression models by interacting with the visual representation of the variable correlations~\cite{klemm20163d, zhang2015visual}, typically a graph representation such as matrices or node-link diagrams.
For interactive numerical prediction, recent work mainly visualizes the quality of regression (e.g., error residuals) to guide users to interactively explore prediction results, select models and tune parameters.
For a detailed description of the pipeline, interaction and tasks of interactive regression, we refer readers to Lu et al.'s recent survey on predictive visual analytics~\cite{lu2017state}.}
\\

\noindent\textbf{Interactive model analysis}. \jiangl{Understanding machine learning models and interpreting their behaviors are important to an informed user interaction in IML applications~\cite{krause2017workflow}.
Hence, interactive model analysis (e.g., understanding and diagnosing models) has attracted recent research attention.
Researchers mainly propose interactive visualizations to facilitate users to reason about the working mechanism of models (e.g., interaction between neurons in deep neural network) and diagnose an unsatisfactory learning process (e.g., convergence failure)~\cite{liu2017towards_survey, liu2018analyzing_noise}.
Liu et al.~\cite{liu2017towards_survey} elaborated on interactive model analysis in a recent survey and we follow their method to categorize the related work.
More recently, Choo and Liu~\cite{choo2018visual} and Hohman et al.~\cite{hohman2018visual} systematically reviewed and summarized visual analytics systems with a focus on the analysis of deep learning.
Readers can find detailed information about interactive model analysis in those previous surveys.}
\section{Research Challenges and Opportunities}
\label{sec:cha_oppt}
Although IML benefits many real-world applications by looping human into the learning process, it still faces several research challenges. 
Here, we highlight the major research challenges, which may pave the way for further research opportunities.
\\

\noindent\textbf{Wide and easy adoption of machine learning techniques}.
Although machine learning provides benefits for almost each industry, its adoption faces several barriers.
First, many learning models perform the same functions (e.g., classification), however, business practitioners usually do not have enough knowledge to decide which one to use. 
As a consequence, it is necessary to study interactive model recommendation, which aims to select the appropriate learning models for a given task and requirement.  
One key challenge here is to leverage interactive visualization to illustrate how the automatic machine learning techniques find the best possible modeling technique for a particular problem. 
Second, parameter settings generally have significant influence on the performance of the learning model.
As a result, a fine tuning of model parameters is required for most of the learning models to obtain satisfactory results.
Accordingly, one challenge is how to systematically explore the multi-dimensional parameter space to greatly reduce the search space of model parameters.
\\

\noindent\textbf{Scalability}.
The size of the model as well as the volume of data are ever-growing in the era of big data.
Visualizing models with complex structures and large numbers of components is challenging, 
especially when the models handle tens of thousands of data items. 
For example, if ResNet-101 is utilized to train a model on a dataset with millions of images and hundreds of categories, it is usually difficult to develop an appropriate visual summarization or abstraction to support model analysis and prediction understanding tasks.
Accordingly, visualization researchers desire to design and develop scalable visualizations to simultaneously convey the model structure, prediction results, and the raw data, as well as to link them together, for a comprehensive analysis of the learning process and results.  
\\

\noindent\textbf{Explainability for data scientists and practitioners}. 
Most existing work focuses on the understanding and analysis of the training process, where the target users are machine learning experts~\cite{liu2017towards_survey}. 
Such type of explainability is very useful for machine learning experts to build robust and high performance learning models. 
However, in real-world settings, data scientists and practitioners also require the explainability of machine learning models to build AI-related applications that business experts can rely on for making informed decisions.
The major obstacle is the trade-off between predictive performance and explainability. 
Typically, decision-tree-based methods can be easily understood by users who are not experts in machine learning. 
On the other hand, ensemble methods such as deep learning models, random forests or boosting trees built upon decisions trees usually achieve better performance.
However, the complexity of the ensemble models usually decreases explainability.
As a result, one interesting venue is to study how to leverage the self-explainable machine learning models such as decision trees to illustrate the working mechanism and predictions of the ensemble models. 
This enables the practitioners to better understand the model predictions and select the appropriate models and parameters for their tasks. 
\\

\noindent\textbf{Monitoring and debugging the online training/testing process of machine learning models}.
In recent two years, there are some initial efforts in debugging offline training process after the training is finished~\cite{choo2018visual}, which already shows the prominence in helping machine learning experts identify the root cause of failure, as well as improve and refine the learning model. 
However, in real-world applications, the training process of many machine learning models is very time-consuming, which can sometimes take up from hours to days~\cite{pu2016variational}.
It is a painful experience to train such models since the training relies on a large number of trial-and-error experiments~\cite{liu2017towards_survey}.
Thus, it is desired to monitor the online training process and find the potential issue(s) that may lead to an unsatisfactory training process.  
One possible solution is to leverage progressive visual analytics techniques to integrate experts into the analysis loop. 
To achieve this, machine learning models are expected to produce semantically meaningful partial results in the training/testing process. 
By leveraging interactive visualizations, experts explore and analyze the system generated partial results, compare newly incoming results with previous ones, a well as perform analysis without having to wait for the entire training process to be completed. 
\\

\noindent\textbf{Improving data quality}.
In machine learning, data and model are equally important.
Thus, data quality is also very important in machine-learning-related applications.
However, during the data collection stage, some inaccurate or error data can be included, which usually affects the further usage of data (e.g. lower the accuracy of the learning models). 
For example, crowdsourcing is a popular way to acquire labeled data. 
Due to individual differences among workers in terms of background, knowledge, and expertise, crowdsourced annotations may be noisy and poor in quality and usually require additional validation~\cite{liu2018interactive}.
As a result, one key issue on preparing and preprocessing data is to ensure the quality and usability of data.
To improve data quality, it is necessary to leveraging interactive visualization for better illustrating the data distribution, identifying potential errors, and minimizing the workload of domain experts who verify the data. 
For example, interactive visualization can be leveraged to tightly combine user-guided methods with system-guided methods during the verification process.

\section{Conclusions}
This paper provides a survey of recent research advances on IML.
We review recent literatures on IML, classify them into a task-oriented taxonomy built by us, and summarize the work focusing on the tasks less discussed by existing surveys, including visual pattern mining, interactive anomaly detection, interactive information retrieval, and visual topic analysis.
We also discuss research opportunities for future work on IML. 
\bibliographystyle{spbasic}      
\bibliography{reference.bib}   


\end{document}